\documentclass[journal]{IEEEtran}

\ifCLASSINFOpdf
\else
\fi

\usepackage{times}
\usepackage{epsfig}
\usepackage{graphicx}
\usepackage{amsmath}
\usepackage{amssymb}

\usepackage{makecell}
\usepackage{multirow}
\newcommand{\conv}[1]{$\left[\begin{array}{ll} \text{1} \times \text{1} \times \text{1} \text{ conv}\\ \text{3} \times \text{3} \times \text{3} \text{ conv} \end{array}\right] \times \text{#1}$}

\newcommand{\cross}[1]{#1 $\times$ #1 $\times$ #1}

\usepackage[pagebackref=true,breaklinks=true,letterpaper=true,colorlinks,bookmarks=false]{hyperref}

\begin{document}

%
\title{3D Aggregated Faster R-CNN for General Lesion Detection}
%
%
%

\author{Ning~Zhang, Yu~Cao,~\IEEEmembership{Senior Member,~IEEE,}
        Benyuan~Liu,~\IEEEmembership{Member,~IEEE,}
        and~Yan~Luo,~\IEEEmembership{Member,~IEEE}
\thanks{}
\thanks{}
\thanks{}}

\maketitle


\begin{abstract}
Lesions are damages and abnormalities in tissues of the human body. Many of them can later turn into fatal diseases such as cancers. Detecting lesions are of great importance for early diagnosis and timely treatment. To this end, Computed Tomography (CT) scans often serve as the screening tool, allowing us to leverage the modern object detection techniques to detect the lesions. However, lesions in CT scans are often small and sparse. The local area of lesions can be very confusing, leading the region based classifier branch of Faster R-CNN easily fail. Therefore, most of the existing state-of-the-art solutions train two types of heterogeneous networks (multi-phase) separately for the candidate generation and the False Positive Reduction (FPR) purposes. In this paper, we enforce an end-to-end 3D Aggregated Faster R-CNN solution by stacking an ``aggregated classifier branch" on the backbone of RPN. This classifier branch is equipped with Feature Aggregation and Local Magnification Layers to enhance the classifier branch. We demonstrate our model can achieve the state of the art performance on both LUNA16 and DeepLesion dataset. Especially, we achieve the best single-model FROC performance on LUNA16 with the inference time being 4.2s per processed scan.
\end{abstract}

\begin{IEEEkeywords}
Medical Image, Small Object Detection, 3D Convolutional Neural Network
\end{IEEEkeywords}

\section{Introduction}
Lesions are damages and abnormalities in tissues of the human body. They can reside in different organs such as livers, bones, lung lobes and even soft tissues. Many of these lesions may develop into cancers. For instance, pulmonary (lung) nodules can grow into lung cancers. Therefore, effective detection of lesions plays an important role in early diagnosis and timely treatment of various cancers. In CT scans, lesions often present distinctive shapes, isolated levels of absorption of X-Ray (Hounsfield Unit value), and other internal structures. All these special visual properties allow both humans and machines to identify and locate lesions. 

However, one challenging aspect is that lesions can be very small and sparse. In contrast, the whole image is massive, eg. a pulmonary nodule can have a diameter smaller than 5$mm$ while a chest CT scan is normally larger than 20,000 $cm^3$ (volume ratio is smaller than 1/100,000). Moreover, positive regions are highly sparse and a typical positive/negative region (anchor) ratio can be $\sim$1/10,000. Note that these issues commonly exist in other medical image modalities such as Diabetic Retinopathy images \cite{chen2019mini} and Hematoxylin and Eosin (H\&E) stained whole-slide images \cite{bejnordi2017diagnostic}. To detect these very small lesions, the false positive rate becomes inevitably high. Therefore, people always have to trade-off between high sensitivity and low false positive rate \cite{setio2017validation}. 

As a result, lesion detectors often contain two major components: (1) candidate generation (region proposal) and (2) false positive reduction (FPR). In the early days, feature-engineering based methods have long been playing an important role. Popular features include Shape Index (SI), Curvedness (CV) \cite{henschke2002ct,murphy2009large,jacobs2014automatic} and other morphology features \cite{setio2017validation}. These features help to indicate suspicious regions. After that, some thresholding procedures are adopted to reduce the false positive rate to strike the balance between recall and precision. Recently, CNN based approaches dominate this task. These CNNs can either be 2D \cite{yan20183d, setio2016pulmonary} or 3D \cite{liao2017evaluate,zhu2018deeplung,zhongliuxie2018}. In general, 3D CNNs often deliver better performance than 2D CNNs. 

Nowadays, the prevalent pipeline of the state of the art solutions \cite{setio2017validation} decouples the two components and trains independent networks for the region proposal component and false positive reduction component respectively. The RPNs are often shallow and fed with a very large 3D input (typically 128$\times$128$\times$128). In contrast, FPRNs are trained with much smaller input (typically 32$\times$32$\times$32, cropped from the suspicious regions detected by RPNs). Thus, FPRNs often employ a deep architecture. Though this two-step pipeline works very well, it is not an end-to-end approach and the whole system becomes very complicated, significantly slowing down the inference speed.



\begin{figure}[t]
    \centering
    \includegraphics[width=1.0\linewidth]{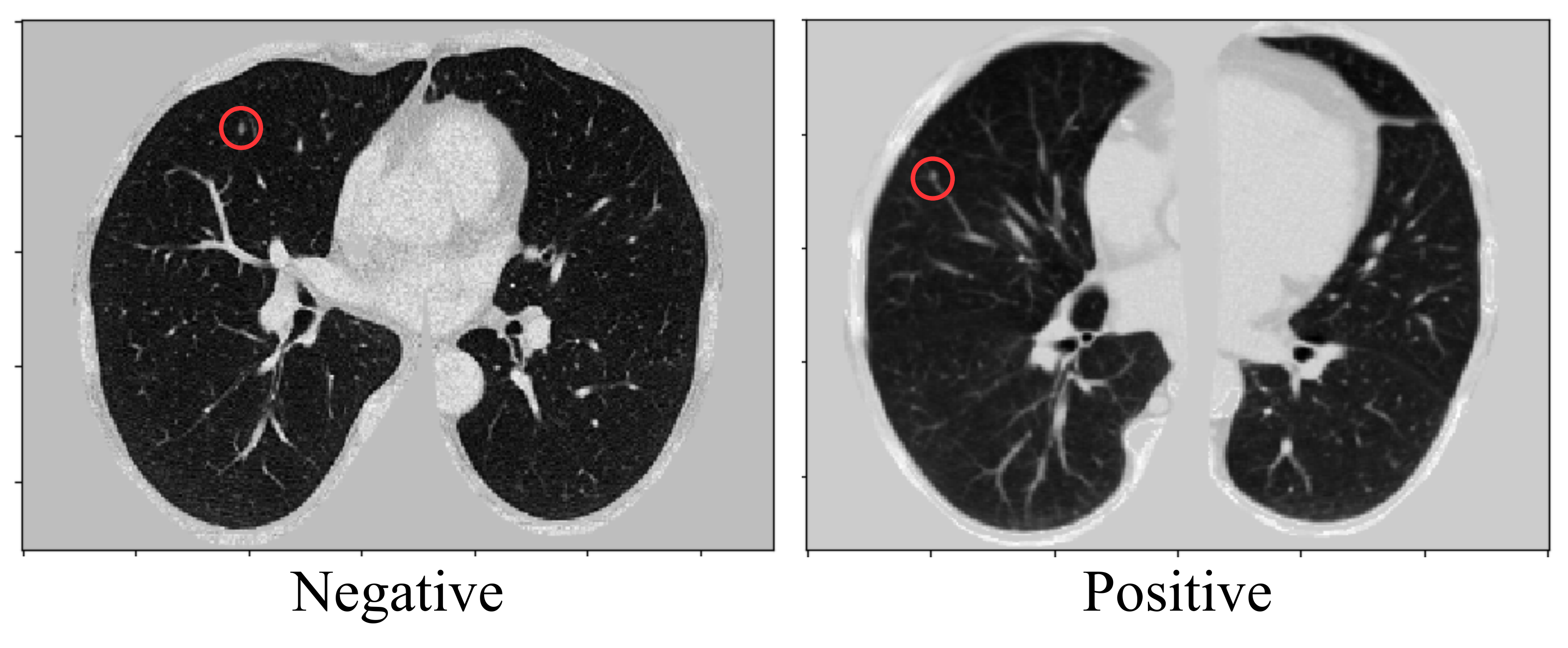}
    \caption{Small positive/negative regions are difficult to distinguish if we only look at the local area (the tiny white round spot). Though both regions appear to be similar, the left one is Negative while the right one is Positive (nodule).}
    \label{small_nodules}
\end{figure}


Motivated by these issues, many end-to-end solutions have been proposed. Xie et. al \cite{zhongliuxie2018} removed the FPR step by fully leveraging a powerful multi-scale 3D RPN. Yan et. al \cite{yan20183d} maintained the balance between high sensitivity and low false positive rate by employing a 2D CNN augmented with 3D context. The 3D context information is gathered by stacking 2D CNN features from neighboring slices. Though they both achieved very good accuracy performance, many drawbacks exist. For instance, in \cite{zhongliuxie2018}, the authors traded off much inference speed for accuracy by adopting very complicated anchor box settings and soft Non-Maximum Suppression (NMS). In \cite{yan20183d}, the 3D object detection task is reduced to a 2D task with the ``center slice" known in advance. However, in real application scenarios, the center slice is always agnostic. As a result, this approach can only detect lesions slice by slice without an efficient retrieval strategy along Z-axis.

The question occurred to us is: why cannot we share the backbone of RPN and FPRN following the idea of Faster R-CNN\cite{ren2015faster}? From our point of view, three major challenges exist. (1) local areas fail to provide sufficient discriminative information to differentiate positive and negative regions (Fig. \ref{small_nodules}). This is especially true for small lesions. (2) objects are too small to apply RoI operations. For instance, in \cite{liao2017evaluate}, the finest resolution of feature layers is 4$mm$ (64 $mm^3$) \footnote{In this paper, we always re-scale images to an isotropic resolution. Therefore, for simplicity, we slightly abuse to use 4 ($mm$) instead of 64 $mm^3$ to represent the voxel resolution.}, while the diameters of a great portion of nodules are smaller than 5mm. In this case, RoI only contains one spatial point at the feature level. Besides, directly increasing the resolution of the feature layer does not help necessarily. (3) prohibitive memory consumption constrains the design of 3D CNNs. We cannot easily adopt a very complicated design for RPNs because of the very large input. On the other hand, FPRNs are often much deeper than RPNs given the much smaller input size. Therefore, directly attaching the FPR branch to the backbone of an RPN may result in poor performance. This depth-memory dilemma also constrains our exploration in the Faster R-CNN direction. 

Our major contribution lies in that we present the first successful approach to enforce an end-to-end full 3D Faster R-CNN solution for the small and sparse lesion detection task. To address the issues and challenges aforementioned: (1) we propose an adaptive Focal Loss to stabilize the training of RPN. This adaptive Focal Loss can well address the large variance introduced from random sub-crops as well as the extremely imbalanced positive/negative anchor ratio. (2) we aggregate multi-scale features to enhance the region-based classifier branch. This feature aggregation enriches the context information in solving the insufficient local area issue. (3) to tackle the failing RoI operation issue and depth-memory dilemma, we magnify the aggregated features locally before feeding them into RoI operations (RoI Align). This local magnification avoids training finer feature layers directly and costs much less memory usage, which allows deeper subset designs for the classifier branch. In the meantime, it also enlarges the RoI area. Even though all our experiments are conducted on 3D CT scans, we argue that our proposed technical components can also be applied to other small and sparse object detection tasks in other image modalities.


\begin{figure*}
\begin{center}
\includegraphics[width=0.9\linewidth]{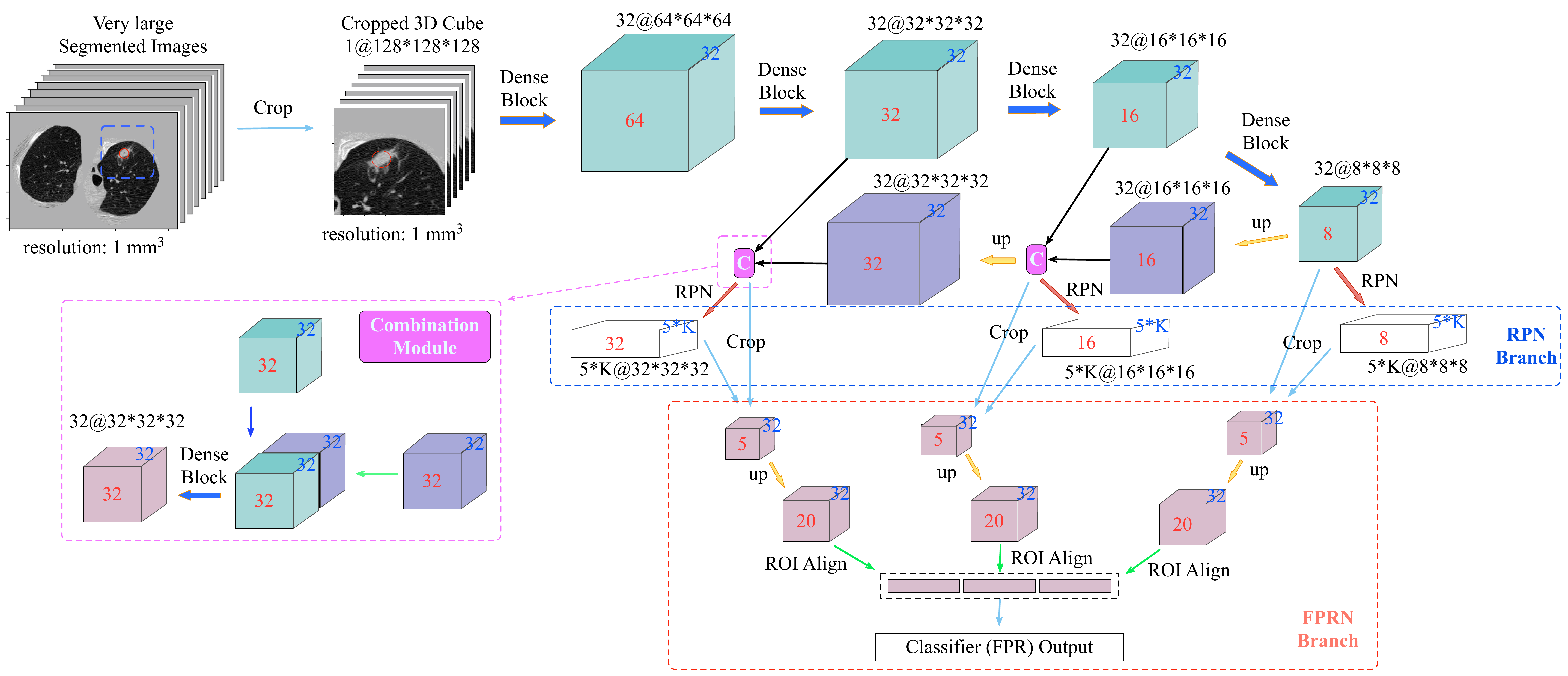}
\end{center}
\caption{The whole network structure with local magnification layers. Sub-cubes are cropped out as the input of a U-Net. We attach an RPN head at each feature scale. Note that we are not sharing parameter across these heads. Proposals from each head would result in aggregated feature crops from all scales instead of a single layer. After RoI operations, these full-scale feature crops would be sent to the ultimate fully connected layers to acquire the FPR score. The final confidence score for each proposal would be the average of two confidence score from both the RPN branch and FPR branch.}
\label{unet_FasterR-CNN}
\end{figure*}

\section{Related Work}
As our work is framed in the object detection task, we will focus on the previous work on object detection and lesion detection. For a more general review about deep learning in the biomedical image domain, one can refer to \cite{mahmud2018applications}.

\textbf{2D Object Detection and Segmentation}. Deconvolutional SSD \cite{fu2017dssd} and Mask R-CNN \cite{he2017mask} are probably the most related solutions in 2D cases. DSSD applies deconvolutions (transposed convolutions) globally on multiple feature scales before the actual detection layers. However, our deconvolutions in Local Magnification operate locally at the RoI crops before RoI operations. This is different from Mask R-CNN, where deconvolutions are located after RoI Align. By doing so, RoI operations can operate on larger areas. In our model, we also adopt RoI Align to reduce the ``Quantization Error" introduced by RoI ops.

In the biomedical image modality, U-Net \cite{ronneberger2015u} was proposed for a segmentation task. Unlike a regular Fully Convolutional Network (FCN) \cite{long2015fully}, U-Net possesses a unique symmetric structure. More importantly, U-Net uses concatenation for lateral connections \cite{he2016deep,Zhang:2017:IMR:3123266.3123332,he2016identity} to enforce the error signal propagation jumping to far early stages. This is different from Feature Pyramid Network (FPN) \cite{lin2017feature} where the element-wise addition is leveraged for the same purpose. 

\textbf{Nodule Detection}. Before the advent of CNNs, people have devised a series of features for this task. Murphy et. al \cite{murphy2009large} proposed to compute Shape Index (SI), and Curvedness (CV) at every position in lungs. Then, a thresholding step finds out seeds lying on the surface of the nodule. After that, a necessary merging process develops individual clusters. These clusters are later formulated as the candidates (Regions of Interest). Some other tailored approaches specialized for certain types of nodules also emerged: for sub-solid nodules \cite{jacobs2014automatic,henschke2002ct}; for large nodules \cite{setio2015automatic}. The main idea is to impose constraints on the Hounsfield Unit (HU) value and the diameter to filter out targeted nodules. Likewise, Tan et. al \cite{tan2011novel} proposed to use three different sets of filters specially designed for 3 different types of nodules: isolated, juxtavascular, and juxtapleural nodules. Note that, nearly all these approaches undergo a re-scaling pre-process step to produce 3D CT images of an isotropic resolution. This re-scaling pre-process is also employed in CNN based approaches.

Recently, many CNN based approaches are gaining increasing attention. Berens et. al proposed a 2D U-Net \cite{ronneberger2015u}. It operates on CT scans slice by slice to determine noisy candidates. These candidates are later merged by morphological analysis. Ding et. al \cite{DBLP:journals/corr/DingLHW17} proposed to fine-tune the models pre-trained with natural images. The idea is to pack 3 consecutive slices of the scan as the RGB channels of a natural image. Ypsulantis et. al \cite{ypsilantis2016recurrent} exploited a recurrent 2D CNN to fully leverage the context information along Z-axis. Compared with plain 2D CNNs, a considerable performance boost is yielded. Concurrently, 3D U-Net variants \cite{zhu2018deeplung,zhongliuxie2018} became popular and achieved huge success. These variants differ in building blocks and the data strategy. Even though all these methods are proposed for pulmonary nodule detection, they can be easily generalized to other types of lesion detection tasks.

\textbf{Divide-and-Conquer}. Though 3D CNNs perform well, one fatal issue is that the input image is too large. As a result, it is not possible to feed the whole image into the whole network (even with the help of semantic segmentation). To address this issue, people resort \cite{liao2017evaluate,zhu2018deeplung,zhongliuxie2018} to the divide-and-conquer strategy: using sub-cubes instead of the whole image as the input. During training, random crops ($\sim$1/6 of the whole image) are fed into the network while at the testing phase, final results are assembled from the detection results from sliding window pieces. 

However, this divide-and-conquer mechanism introduces too much randomness during training, slowing down the convergence speed. One straight forward explanation comes from the batch normalization \cite{ioffe2015batch} layers. Batch normalization focuses on channel-wise whole feature map statistics while with sub-cubes being fed in, these statistics become unstable. Another reason lies in the commonly used online hard negative mining mechanism \cite{shrivastava2016training}. In each iteration, only the hard negatives contribute to the loss. However, these hard negatives are highly varying across iterations. This large variance further affect the convergence speed. One quick remedy can be the focal loss \cite{lin2017focal} as it takes into account all samples when calculating the loss. However, the vanilla form of focal suffers from the extremely small positive/negative sample ratio. Our adaptive focal loss is motivated by these observations. 

\textbf{False Positive Reduction}. At this stage, RoIs are assumed to be ready to extract 3D cubes to train independent CNNs. Setio et. al \cite{setio2016pulmonary} proposed a 2D Multi-View CNN for this task. The 3D context is encoded by 9 different plains of symmetry extracted from each candidate cube. These 2D plains are then fed into 2D CNNs. CNN features are fused to make the final decisions. Dou et. al \cite{dou2017multilevel} devised 3 shallow but powerful 3D CNNs to ensemble for this task. Each of these 3 CNNs tackles nodules of different sizes. Some other 3D CNNs also reportedly work well such as 3D U-Net CNN (PAtech Team) and 3D Wide Residual Network \cite{zagoruyko2016wide}. In general, compared with 2D CNNs, 3D CNNs seem to work better with this task, since it is the most straight forward way to leverage the power of CNNs.

\textbf{Full Solutions}. Currently, multi-phase (ensemble) solutions outperform single-phase (single-model) ones. The detection Network and the FPR network are often decoupled and specialized independently \cite{setio2017validation}. In other words, the two networks do not share the backbone, which makes the whole solution complicated and systematically slow. Most of these solutions adopt 3D CNNs for both types of networks. However, it is reported in \cite{yan20183d} that the two-step pipeline with 3D CNNs may fail in much noisy settings in terms of both image quality and less precise annotations (ex. DeepLesion dataset \cite{Yan_2018_CVPR}). Yan adopts a 2D CNN enhanced with multiple neighboring slices context to attack the issue and better performance is acquired. 
\begin{figure*}[t]
\begin{center}
\includegraphics[width=0.95\linewidth]{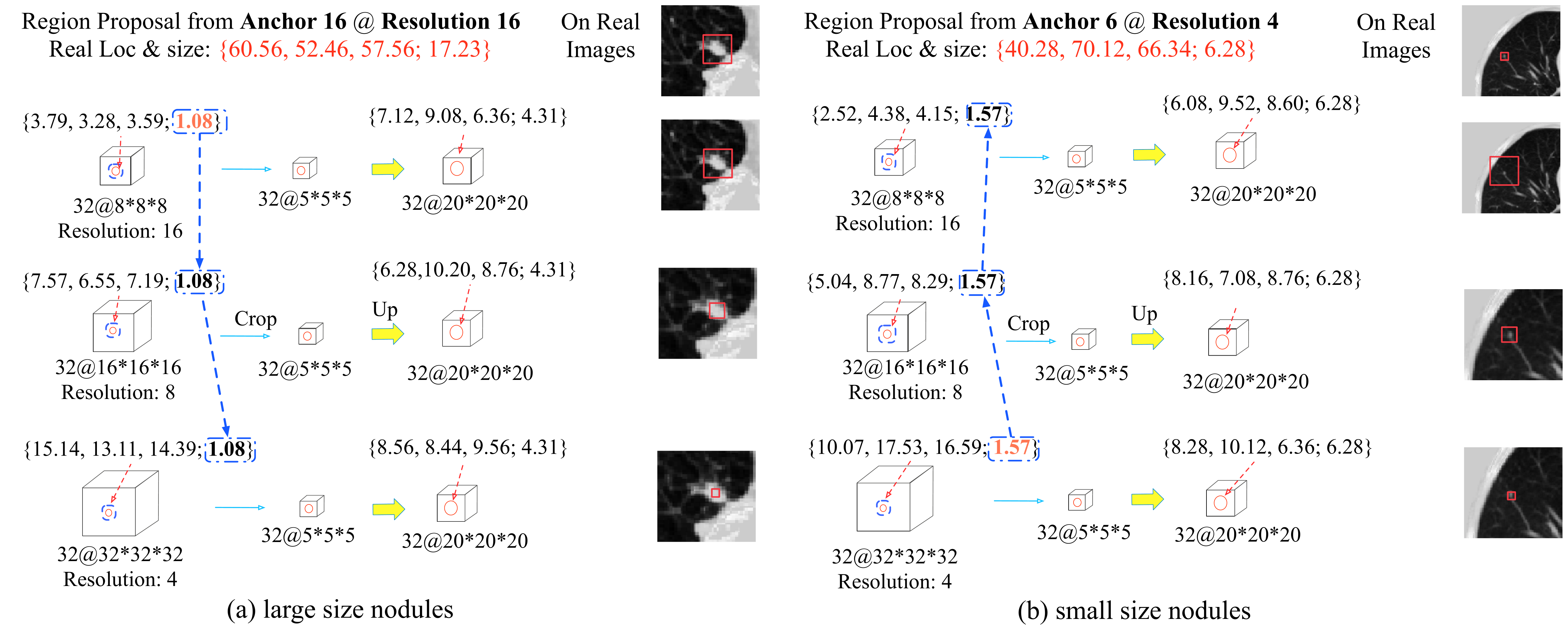}
\end{center}
\caption{Diameter Alignment and Local Magnification operations in the FPRN branch. In our case, ignoring the confidence score, each proposal is represented as a 4-dimension vector: \{Z, Y, X, Diameter\}. Any proposal at each scale will require an all-scale feature crop. RoI spatial locations are calculated following the standard routines \cite{ren2015faster}, while the crop size at the referenced level (where the proposal is acquired) will be broadcasted to other levels of the pyramid. For instance, in (a), the proposal \{60.56, 52.46, 57.56; 17.23\} is derived from ``Resolution 16" making the reference scale be 16 and the diameter be 1.08. Then the diameter 1.08 would be top-downed to other levels. Following the same rule, in (b), the diameter would be bottom-upped from ``Resolution 4" to the other two scales. All these crops would be up-sampled before they are aggregated. Note that we crop the feature maps with some margins with a fixed crop size.}
\label{FPR_Crop}
\end{figure*}

\section{Our Approach}
As illustrated in Fig. \ref{unet_FasterR-CNN}, our model contains two heads at each pyramid level: Region Proposal Network (RPN) and False Positive Reduction Network (FPRN). They share the same backbone of a U-Net structure with DenseNet building blocks \cite{huang2017densely}. This design allows the end-to-end training.

\subsection{Backbone Network}
Our backbone network employs a U-Net structure. In the upstream pathway, feature map sizes are gradually reduced to extract increasingly abstract features, while in the down-stream pathway, upsampling operations (transposed convolutions) take effect to acquire information complement to the upstream pathway. Note that feature maps of the same level from two pathways are concatenated (the pink Combination Module in Fig. \ref{unet_FasterR-CNN}) before they propagate to the next layer. 

This feature pyramid idea is also explored in Feature Pyramid Network (FPN) \cite{lin2017feature} where ``element-wise addition" is used instead of ``concatenation" for the lateral shortcut. This mental image  of ``addition" vs. ``concatenation" reminds us of the difference between ResNet and DenseNet: skip connections \cite{he2016deep,Zhang:2017:IMR:3123266.3123332,he2016identity} vs. concatenations. For this reason, we use DenseNet Building Block in our model. Besides, in FPN, up-sampling operations are parameter-free (using interpolations) while we adopt transposed convolutions which introduce some additional free parameters.

All detailed layer configurations are shown in Table \ref{U-DenseFPR}. We use the same notations in \cite{huang2017densely}. 
Note that, we configure all feature maps in Feature Pyramid to contain 32 channels, making it convenient for head sharing across the RPN branches (in practice, we actually do not share the RPN heads) as well as for the feature aggregation in the FPRN branches.

\begin{table*}[t]
\caption{Layer Configurations. ``max" and ``avg" denote Max Pooling and Avg Pooling respectively. The growth rate is set to be 16. Following the same notations from DenseNet most of ``conv'' layers shown in the table correspond the sequence BN-ReLU-Conv.} 
\centering
\resizebox{0.95\textwidth}{!}{%
\begin{tabular}{c|c|c|c|c|c|c|c}
\hline
Encoding & PreBlock & Encode (1) & Transition (1) & Encode (2) & Transition (2) & Encode (3) & Transition (3)  \\ \hline
Layers & $\begin{array}{ll} \text{3} \times \text{3}\times \text{3} \text{ conv}\\ \text{3}\times \text{3}\times \text{3} \text{ conv} \\ \text{2}\times \text{2}\times \text{2} \text{ max} \end{array} $
&\multicolumn{1}{l|}{\conv{2}}  & $\begin{array}{ll} \text{1}\times\text{1}\times \text{1} \text{ conv}\\ \text{2}\times\text{2}\times \text{2} \text{ avg} \end{array}$ 
&\multicolumn{1}{l|}{\conv{2}}   & $\begin{array}{ll} \text{1}\times\text{1}\times \text{1} \text{ conv}\\ \text{2}\times\text{2}\times \text{2} \text{ avg} \end{array} $
&\multicolumn{1}{l|}{\conv{2}}   & $\begin{array}{ll} \text{1}\times\text{1}\times \text{1} \text{ conv}\\ \text{2}\times\text{2}\times \text{2} \text{ avg} \end{array}$ \\ \hline
Output & \cross{64} & \cross{64} & \cross{32} & \cross{32} & \cross{16} & \cross{16} & \cross{8}\\ \Xhline{3\arrayrulewidth}
Decoding & Detector & Transition (1) &Decode (1) & Upsample (2) & Transition (2) & Decode (2) & Upsample (3)  \\ \hline
Layers &$\begin{array}{ccc} \text{Bottle Neck} \times 2\\ \text{1}\times \text{1}\times \text{1} \text{ conv} \end{array}$
& $\text{1} \times \text{1}\times \text{1} \text{ conv}$
&\multicolumn{1}{l|}{\conv{2}} 
& $\begin{array}{ccc} \text{2}\times \text{2}\times \text{2} \text{ deconv}\\ \text{concat} \\ \text{1}\times \text{1}\times \text{1} \text{ conv} \end{array}$
& $\text{1} \times \text{1}\times \text{1} \text{ conv}$&\multicolumn{1}{l|}{\conv{2}} 
& $\begin{array}{ccc} \text{2}\times \text{2}\times \text{2} \text{ deconv}\\ \text{concat} \\ \text{1} \times \text{1} \times \text{1} \text{ conv} \end{array} $\\\hline
Output & * & \cross{32} & \cross{32} & \cross{32} & \cross{16} & \cross{16} & \cross{16} \\
\Xhline{3\arrayrulewidth}
Magnify & $\left[\begin{array}{l}\text{2} \times \text{2} \times \text{2} \text{ deconv} \\ \text{Bottle Neck} \times 2 \end{array}\right] \times \text{2} $
& Magnify Output & \cross{5} & RoI Align Output & \cross{2}  & Classifier & $\begin{array}{ll}
     & \text{768} \times \text{256} \text{ full} \\
     & \text{256} \times \text{ 1 } \text{ full} 
\end{array}  $
\\ \hline

\end{tabular}
}
\vspace{1 ex}

\label{U-DenseFPR}
\vspace{-3 ex}
\end{table*}

\subsection{Region Proposal Network Branch}
We explore multi-scale techniques in our model to better handle the large variance of the object size. \cite{lin2014microsoft} suggests that sharing heads across different pyramid levels can result in some performance boost. However, we observed that branches keeping independent to each other is a better choice in our case. Unlike the 2D object detection where bounding boxes are represented as 4-element vectors: \{x, y, w, h\}, we use \{z, y, x, diameter\} to encode a bounding box here. This is because that in field practice, radiologists and physicians adopt this way to annotate lesions. 

Distributing anchors to different scales is important here to allow ``fair hits" for nodules of different diameters. ``Fair hits" means nodules of different diameters have similar hit counts on box templates. It helps to balance sparse positive samples and avoiding small nodules being flushed out by large nodules. Another good attribute brought by this technique is that Regions of Interest would have a similar size on their respective reference pyramid levels (Fig \ref{FPR_Crop}). For instance, a lesion of 6mm diameter on stride 4mm feature maps and a lesion of 12mm on stride 8mm feature maps would have the same size. 

The training loss of RPN branches contains two parts: Bounding Box Regression and Binary Classification. For the Bounding Box Regression part, the standard Smooth L1 Loss is adopted. To the classification end, two popular options are Online Hard Example Mining (OHEM) and Focal Loss \cite{lin2017focal}, which is defined as:
\begin{equation}
    \mathcal{L}_F(p_t) = -\alpha_t(1-p_t)^\gamma log(p_t),
\label{vfocal_sub}
\end{equation}

\begin{equation}
    \mathcal{L} = 1/N_{pos} \sum_{i}^{N_{all}}{L}_F^{i},
\label{vfocal}
\end{equation}
where $p_t^{i} = (p^{i})^y(1-p^{i})^{(1-y)}$, $p$, $y$ denote the output probability and the ground truth respectively. $\gamma$ and $\alpha$ are the hyperparameters to control the ``loss decay" rate and ratio between losses from positive and negative samples. 

In our case, the divide-and-conquer strategy is introduced, resulting in a large batch-wise variance across training iterations. This large variance leads OHEM to be unstable because only a few anchors contribute to the loss. In contrast, Focal Loss \cite{lin2017focal} covering all anchors should stabilize the training. However, the vanilla form (defined as Eq. \ref{vfocal_sub} and \ref{vfocal}) does not work well in our extremely small and sparse object detection task. The reason lies in the denominator in Eq. \ref{vfocal}. In \cite{lin2017focal}, it is determined by the number of positive samples. In our case, the positive/negative ratio is extremely small ($\sim$1/10000). Therefore, using $N_{pos}$ as the denominator does not work here. Instead of tuning the denominator directly, we adjust the loss function as follows:
\begin{equation}
\begin{aligned}
    \mathcal{L} &= T log(N_{TN})/N_{TN}\sum_{N_{neg}}\mathcal{L}_F(p_t) \\
    &+ 1/N_{pos}\sum_{N_{pos}}\mathcal{L}_{CE}(p_{pos}),
\end{aligned}
\label{tfocal}
\end{equation}
where $N_{TN}$, $N_{pos}$, $N_{neg}$ denote respectively the number of True Negative samples, positive samples and negative samples. $T$ is a linear factor increasing with iterations. 

By doing so, we introduce a ``focus shift" throughout the whole training. At the initial training phase, we place the focus on the massive negative samples because this dense updating signal should move more ``safely" towards convergence. As the training proceeds, the focus shifts to positive samples to avoid ``overkill". In practice, the term $1/N_{TN}\sum_{N_{neg}}\mathcal{L}_F(p_t)$ would drop very fast (much faster than quadratically). Therefore, we add the linear and log multiplier to smooth out this ``focus shift" process. Note that we use Focal Loss only for negative samples while we calculate Cross Entropy for positive samples, given the small number of positive samples. 


\subsection{False Positive Reduction Network Branch}
We adopt an aggregated classifier in the FPRN branch in our model. Unlike in common practice \cite{ren2015faster, lin2017feature} that feature crops only come the single reference scale features, each proposal in our model will result in a feature aggregation across the feature pyramid. This feature aggregation is realized by Diameter Alignment (Fig. \ref{FPR_Crop}). A good attribute of this aggregated classifier is that we can explicitly enforce the scale-invariant property for nodules of different sizes by sharing the weights of the classifier heads.

Diameter Alignment (DA) (Fig. \ref{FPR_Crop}) helps to incorporate more context information for small nodules and to probe into in-nodule details for large nodules. For middle size nodules, both context and in-nodule detail information is enhanced. The core idea is that the feature crop size at a certain scale of the feature pyramid will be broadcasted to other scales with the centroid remaining the same. The resulting aggregated feature would automatically incorporate more context information. Note that in our implementation, we crop with some margins to ensure the transposed convolution in later Local Magnification works correctly.

This context information enrichment is motivated by the observation that if we only look at local areas (RoI), blood vessels and small nodules can appear quite similar to each other as small white spots (Fig. \ref{small_nodules}). To handle this issue, in real clinical practice, radiologists often scroll up and down along Z-axis to examine the context when detecting confusing small nodules. We show in later ablation study that this Diameter Alignment plays an important role in improving the FPRN branch. 

We adopt RoI Align \cite{he2017mask} in our models as it can theoretically work with small regions. We follow the official implementation of RoI Align and extend it to the 3D case (from 4 neighboring points to 8 neighbor points). All RoI Align outputs will be aggregated (by concatenation) before they reach the fully connected layers. The final confidence score of each proposal will be the average of two branches. 

\subsection{Local Magnification}
Finer resolution of feature layers is another option to attack the small RoI issue. However, naive approaches can easily fail because of the prohibitive memory consumption. To circumvent this problem, we propose to up-sample RoI crops locally to avoid the tremendous memory consumption. Conceptually, this operation provides a ``closer" look at the regions of interest (Fig. \ref{FPR_Crop}), which resembles much putting a magnifying lens above RoIs. Therefore, we name this operation as Local Magnification.

\subsection{Joint Training vs. Alternating Training Between Branches}
Essentially the whole model contains two branches: RPN and FPRN. Since we have no ready-to-use 3D model as for the 2D scenarios on which we can directly fine-tune, we have to train the model from scratch. We first train the RPN first to acquire a good initialization for the backbone given the ``dense" error signal emitted by the RPN branch. After that, we add the FPRN branch to the model.

After we stack the FPRN branch on the model, we have multiple options for further training: joint training and alternating training between branches. In practice, we found that the former choice seems to be more stable in terms of performance. We attribute this fact to the combination of both ``Global and Local" complementary losses. In a way, RPN losses are more general and global which take into account all spatial positions while FPRN losses only focus on the highly suspicious regions, making them more local.

\begin{table*}
\caption{Ablation study for baseline RPNs, we use DenseNet Block as the building block for both the upstream and downstream pathway. The detailed architecture of the model can be found in Table \ref{U-DenseFPR}. Note that we do not share the RPN heads. Each pyramid level has its own RPN head of an identical structure. ``$\dagger$" means using the ``Anchor Based Sampling" technique, which can be interpreted as a boosting procedure. With ABS, an additional 30 epochs of training is conducted.}
    \centering
    \begin{tabular}{c|c|c|c|c} 
    Backbone & Multi-Scale & OHEM & Focal Loss & FROC/Sensitivity \\ \hline
Res18 \cite{zhu2018deeplung}& & \checkmark & &0.834 / 0.946 \\ \hline
DualPath \cite{zhu2018deeplung}& & \checkmark & &0.842 / 0.958 \\ \hline
ResBlock \cite{zhongliuxie2018} &\checkmark & \checkmark & & \textbf{0.920} / - \\ \hline
ResBlock$\dagger$ \cite{zhongliuxie2018} &\checkmark & \checkmark & &\textbf{0.935} / - \\ \Xhline{3\arrayrulewidth}
a. DenseBlock + anchor1 & & \checkmark & & 0.839 / 0.896 \\ \hline
b. DenseBlock + anchor1 & &  & \checkmark & 0.868 / 0.941 \\ \hline
c. DenseBlock + anchor1 &\checkmark & \checkmark &  & 0.898 / 0.956 \\ \hline
d. DenseBlock + anchor1 &\checkmark &  & \checkmark &\textbf{0.910} / \textbf{0.982} \\ \hline
e. DenseBlock + anchor2 &\checkmark &  & \checkmark & 0.899 / 0.970 \\ \Xhline{3\arrayrulewidth}
f. DenseBlock + anchor1$\dagger$ &\checkmark &  & \checkmark & \textbf{0.917} / 0.977 \\ \hline
 \end{tabular}

\label{tb:baseline_results}
\end{table*} 

\section{Experiment on LUNA16}
We conduct a series of experiments on the task of Pulmonary Nodule detection with the LUNA16 dataset \cite{setio2017validation}. This dataset is a subset of the publicly available dataset LIDC-IDRI \cite{armato2011lung}. It summarizes the annotations from LIDC-IDRI: tiny nodules (diameter $<$ 3mm) and annotations of low confidence (fewer than 2 physicians agree on) are excluded. As a result, LUNA16 contains 888 CT scans and 1186 nodules. We follow the rule of the LUNA16 challenge \cite{setio2017validation} by conducting 10-fold cross validation and evaluate the performance with the official FROC score: the average recall rate with the number of false positives being 0.125, 0.25, 0.5, 1, 2, 4, 8 per scan. In LUNA16 Challenge, a 3D region proposal is counted as a True Positive as long as its center is located inside a true bounding sphere: the distance between two centers is less than the radius of true bounding sphere. We adopt this criterion for all our experiments, including the experiments on the DeepLesion dataset.

\subsection{Data Preparation}
As in \cite{liao2017evaluate}, we use officially provided segmentation masks to remove unnecessary volume from the original 3D scans. We rescale all images to $1mm \times 1mm \times 1mm$. During training, we randomly crop out 3D $128\times128\times128$ cubes from the pre-processed images as the input. Note that the segmented scans are typically much larger (6 $\sim$ 8 times). During test, we adopt the sliding window style cropping: divide the whole images into overlapped cubes and merge all detection results later for the final decision.

\subsection{Baseline RPNs}
We adopt a DenseNet backbone. We conduct extensive ablation experiments to evaluate the effectiveness of all the technical components. To make fair comparisons, all ablation experiments undergo a 10-fold cross validation with 50 training epochs. During training, IoU thresholds are $IoU_+=0.5$ and $IoU_-=0.02$. Regions possessing IoU value with Ground Truth larger than $IoU_{+}$, smaller than $IoU_{-}$ and in between will be assigned as positive, negative and ignored respectively. The $\alpha$ and $\gamma$ in the Focal Loss are set to be 0.8 and 5 respectively. 

In the ablation study, when multi-scale technique is removed, anchors (denoted as ``anchor1") are set to be \{4, 6, 8, 12, 16, 24\}mm at the feature layer of 4mm resolution, while with the multi-scale technique the \{4, 6\}mm, \{8, 12\}mm \{16, 24\}mm anchors are distributed to the feature level of 4mm, 8mm, 16mm resolution respectively. As Eggert et al. \cite{eggert2017closer} suggest that the anchor set matters much for small object detection tasks, we also explore other anchor sets such as \{5, 10\},\{15, 20\}, \{25, 35\} (denoted as ``anchor2") to validate this phenomenon. All results are summarized in Table \ref{tb:baseline_results}.

\textbf{OHEM vs. Focal Loss}. Our proposed Focal Loss always results in a large performance improvement compared with OHEM (and usually use fewer iterations). This can be clearly illustrated by comparing Row a and Row b (0.839/0.896 $\rightarrow$ 0.868/0.941), Row c and Row d (0.898/0.956 $\rightarrow$ 0.910/0.982). Note that we do not report the experiment result with the vanilla Focal Loss because it simply does not work.

\textbf{Multiscale vs. Single Scale}. Leveraging multi-scale technique drastically improves the performance (over 0.05 FROC score improvement). This is well demonstrated by comparing Row a and Row c (0.839/0.896 $\rightarrow$ 0.898/0.956), Row b and Row c (0.868/0.941 $\rightarrow$ 0.910/0.982).


\textbf{DenseBlock vs ResBlock}. So far, we have set up a relatively strong baseline network. However, we cannot directly compare our Dense Block with Res Block or with DualPath Block. The reason is two-fold. First, our model architecture may not be the best. We do not follow the common practice to increase the kernel number gradually as the feature map gets smaller. This is due to the idea that we want to force the RPN heads to be of identical shape, which makes it convenient to share heads and to build an aggregation of features afterward. Second, there are many other factors that also affect the performance such as re-shuffle and cropping strategies, hyper-parameter settings and the implementation details. This is illustrated perfectly by the large performance gap between the two Res18-like networks \cite{zhu2018deeplung,zhongliuxie2018}. Effective modifications include substituting the ReLU activation and NMS to Randomized ReLU and soft-NMS respectively as well as leveraging the multi-scale technique.

\subsection{Anchor Based Sampling}
Xie et al. \cite{zhongliuxie2018} have shown that ``Anchor Based Sampling" (ABS) can greatly improve the performance. This ABS works by ``boosting", i.e. focusing more on ``hard cases" which easily confuse the model. This mechanism functions by repeating the following steps: (1) training as usual for certain iterations; (2) testing each whole CT scan in the training set to locate the hard regions. These hard regions will be more focused in the next training round. We apply one round ABS in our models and find it work considerably well. Results are summarized in Table \ref{tb:baseline_results} and Table \ref{tb:fpr_results}.

\begin{table*}
\caption{Ablation study on the FPRN branch (FROC/Sensitivity). ``Crop 96" and ``Crop 128" represent the input cube size be 96 and 128 respectively. ``DA" is short for Diameter Alignment. If it is removed, we still have full-scale feature aggregations. However, we will use the standard procedure to calculate nodule diameters across all pyramid levels. ``Magnify", ``Joint" and ``Alternating" respectively denote Local Magnification ops, training the RPN and FPRN branches simultaneously and alternatively. We also report the ensemble performance of the 3D Aggregated Faster R-CNN with and without Local Magnification. We merge the results by simply averaging the overlapped proposals. Non-overlapped proposals between the two proposal sets are retained (Union) or discarded (Intersection).}
    \centering
    \begin{tabular}{c|c|c|c|c|c|c|c} 
    Model & DA & Magnify & Joint & Alternating & RPN & FPRN & Combined\\ \hline
Crop 96 &\checkmark &  & \checkmark & & 0.911/0.970& 0.889/0.946 &0.917/0.970\\ \hline
Crop 96 &\checkmark & \checkmark & \checkmark  & &\textbf{0.916/0.977} & \textbf{0.902/0.963} & \textbf{0.919/0.976} \\ \hline
Crop 96 &\checkmark & \checkmark &  &\checkmark& 0.908/0.972 &0.894/0.962 & 0.912/0.972 \\ \Xhline{3\arrayrulewidth}  
Crop 128 &\checkmark &  & \checkmark & &0.923/0.976 &0.889/0.944 & 0.926/0.976\\ \hline
Crop 128 &\checkmark & \checkmark & \checkmark &   & \textbf{0.925/0.983} & \textbf{0.905/0.957} & \textbf{0.930/0.981} \\ \Xhline{3\arrayrulewidth} 
Crop 128$\dagger$ & &  & \checkmark &   & 0.920/0.972  & 0.895/0.955  & 0.926/0.964 \\ \hline
Crop 128$\dagger$ &\checkmark &  & \checkmark & &\textbf{0.930/0.983} &0.908/0.963 & \textbf{0.939/0.985}\\ \hline
Crop 128$\dagger$ &\checkmark & \checkmark & \checkmark &   & 0.928/0.984  & \textbf{0.914/0.968 } & 0.935/0.983 \\ \hline
Merged (Intersection)&\checkmark & \checkmark & \checkmark &   & \textbf{- } & \textbf{-} & \textbf{0.943}/0.979  \\ \hline
Merged (Union) &\checkmark & \checkmark & \checkmark &   & \textbf{- } & \textbf{- } & 0.942/\textbf{0.991} \\ \hline
 \end{tabular}

\label{tb:fpr_results}
\end{table*}

\begin{table}
\caption{The state of the art single-model solutions. Note that our inference time is evaluated with much inferior GPU settings compared with \cite{zhongliuxie2018}}
    \centering
    \begin{tabular}{c|c|c} 
    Model & FROC & Inference Time\\ \hline
    Res18 \cite{zhu2018deeplung} & 0.834 & -\\\hline
    DualPath \cite{zhu2018deeplung} & 0.842 & - \\\hline
    ResBlock$\dagger$ \cite{zhongliuxie2018} & 0.935 & 15s /Scan\\ \hline
    3D-AG (Ours) & \textbf{0.939} & \textbf{4.2s /Scan}\\ \hline
 \end{tabular}
 
\label{tb:stateofart}
\end{table}

\subsection{False Positive Reduction Network Branch}
We stack the FPRN branch over the backbone of RPN and use joint (multi-task) training for the whole model. Unlike the regular Faster R-CNN classifier branch, three main modifications are introduced in our FPRN branch: (1) multi-scale Feature Aggregation, (2) Diameter Alignment when cropping RoIs, (3) Local Magnification. Note that all Local Magnification layers' weights are shared across all pyramid levels. We conduct a full ablation study to isolate the effect brought by each technique. Furthermore, we also compare the effects of the input crop size and the two different training strategies: training two branches jointly and alternating between the training of two branches. We refer to our model with and without Local Magnification as ``3D-AG" and ``3D-AG-LM". Results are summarized in Table \ref{tb:fpr_results}. 

As we can see from Table \ref{tb:fpr_results}: (1) introducing FPR branch brings in consistent and considerable performance boost (0.917/0.977 $\rightarrow$ 0.939/0.985). To our surprise, it also improves the RPN branch (0.917/0.977 $\rightarrow$ 0.930/0.983). We attribute this to the strong scale-invariant constraints imposed by the FPRN branch during training, forcing the backbone network responding to both branches simultaneously. This conjecture is further supported by the fact that the joint training performs better than the alternative training. (2) local magnification layers bring in a consistent positive effect for the FPRN branch in terms of both FROC score and Sensitivity (0.895/0.955 $\rightarrow$ 0.914/0.968). However, when we reach the final result, models with Local Magnification sometimes trail ones without it. We leave this inconsistency as our future work. (3) Diameter Alignment is critical (0.926/0.964 $\rightarrow$ 0.939/0.985). Once we remove it, RPN fails to improve compared with the baseline RPN (0.920/0.972 vs. 0.917/0.977) and the FPRN branch drastically loses efficacy (from 0.908/0.963 to 0.895/0.955).


\subsection{LUNA16 Leader Board}
Our models achieve the best FROC score among single-model solutions \ref{tb:stateofart}. However, recently many better results from multi-phase (ensemble) solutions have been reported. Unfortunately, important details are still missing. To our best knowledge, all these state-of-the-art ensemble solutions consist of multiple heterogeneous networks (ensembles) such as PAtech (0.951, 1 RPN + 2 FPRNs); JianPeiCAD (0.950, 2 RPNs + 1 FPRN ); LUNA16FONOVACAD (0.947, 1 RPN + 3 FPRNs) \footnote{Results can be found on https://luna16.grand-challenge.org/results/}. Our Merged (union) and Single-Model results would rank at 4th and 6th place on the LUNA16 Leader Board. Nevertheless, our model still works considerably well given the fact that our models allow end-to-end training and are inherently faster than ensemble solutions. 

 

\begin{figure*}[t]
\begin{center}
\includegraphics[width=1.0\linewidth]{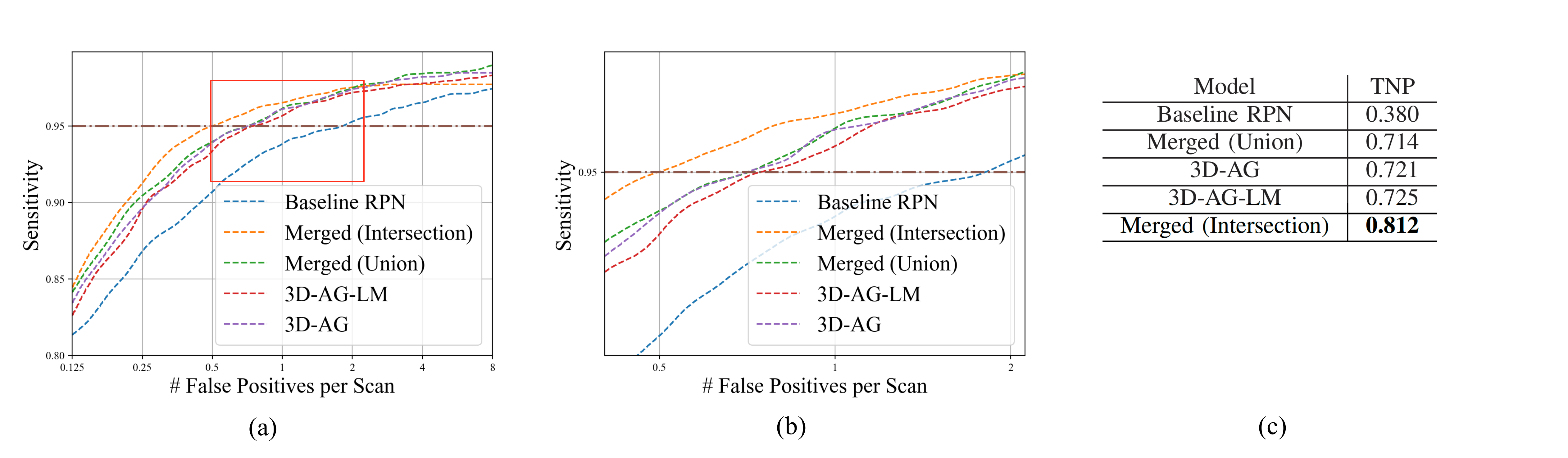}
\end{center}
\caption{FROC Curve and True Negative Patient Scores of our models. (a) presents the original FROC evaluation curves. (b) shows detailed information when all curves hit the 0.95 sensitivity. (c) summarizes the TNP score of each model.}
\label{bootstrapping}
\end{figure*} 
\begin{figure*}
\begin{center}
\includegraphics[width=1.0\linewidth]{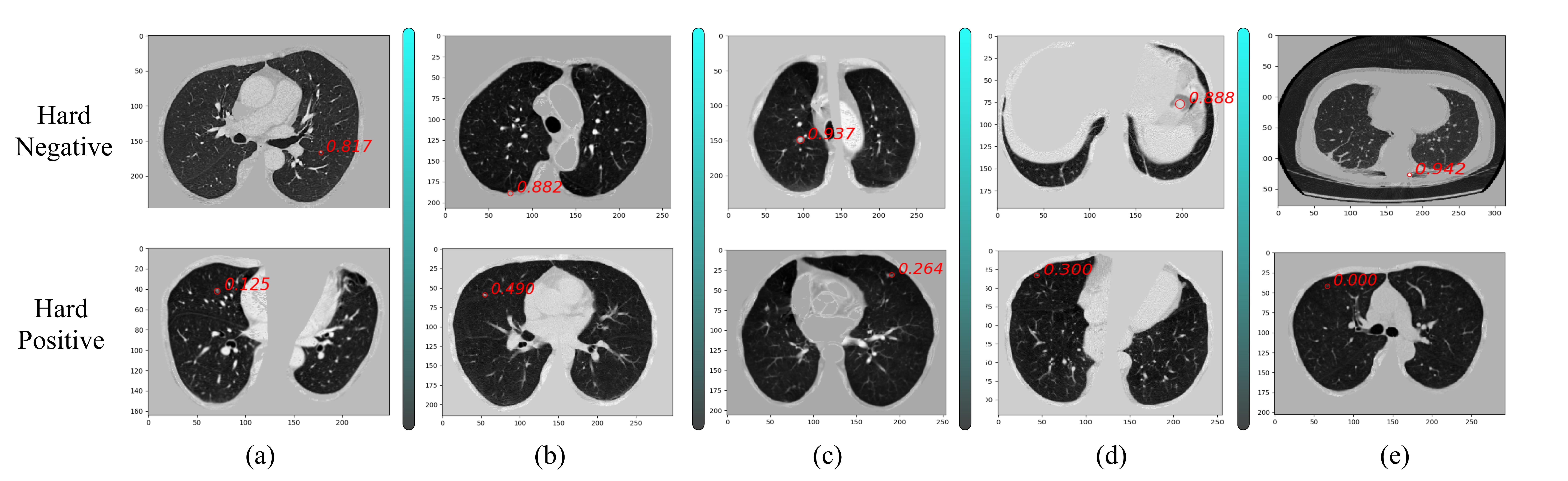}
\end{center}
\caption{Typical Hard Cases confusing our models. We use the ``Faster R-CNN w/ Magnify" as the testing model. The probability of each region is marked on the images. Other models share a similar property. The score of 0 indicates the nodule is missed.}
\label{hard_cases}
\end{figure*} 

\subsection{Inference Time}
We test our models with 4 Tesla K80 GPUs (48 GB) with an Intel(R) Xeon(R) E5-2640 v2 CPU. We set the probability threshold to be 0.269 (Sigmoid(-1)). The inference time for baseline RPN models and 3D Aggregated Faster R-CNN models without and with Local Magnification are $\sim$3.0s, $\sim$4.2s and $\sim$5.0s per pre-processed scan respectively. In \cite{zhongliuxie2018}, 4 Titan XPs (48 GB) are adopted which are much faster than Tesla K80 (12.1 TFLOPs vs. 5.6 TFLOPs). Despite this inferior GPU setting, our approaches are at least 2 times faster than \cite{zhongliuxie2018} ($\sim$15s per pre-processed scan). 


Note that detailed information about the inference time of other state-of-the-art ensemble approaches is unavailable to the public. Nevertheless, we argue that our approach is systematically faster. Because we cut off the time for additional cropped cubes (raw images) to propagate through very deep (much deeper than the RPN backbone) 3D CNNs. 

\subsection{True Negative Patients}
Though the FROC score is well designed for this task, it only focuses on the individual lesion level. We argue that more attention needs to be paid to at the Patient level since it is more acceptable for Positive Patient to have False Positives (FP) than for Negative Patients. To this end, we also report the performance of our models on the True Negative Patient (TNP) score $N_{TNP}/N_{NP}$.

Based on LUNA16 annotations, there are 287 negative patients. We adopt the probability thresholds allowing the sensitivity to reach 0.95 and calculate the TNP score for each model. As shown in Fig. \ref{bootstrapping}, a higher FROC score does not necessarily associate with a better TNP score. For instance, ``Merged (Union)" processes much higher FROC score than all single-model solutions while yielding a worse TNP score. This is also another reason that we argue TNP evaluation is an important complement to FROC evaluation. 

\subsection{Visualization of Hard Cases}
We visualize hard cases in Fig. \ref{hard_cases}, including positives hard to detect and negatives easy to mistake. Typically, False Positives (Hard Negatives) are caused by small nodules (a-c), the failure of segmentation (d), the noise and bad quality of raw images (e). As for False Negatives (Hard Positives), the model may additionally suffer from the low contrast of RoIs with the background (d, e). 
Therefore, we argue that better segmentation and higher quality of raw CT scans should further help the detection. 

\section{Experiments on DeepLesion}
We also evaluate our model on a more general lesion detection task with the DeepLesion \cite{Yan_2018_CVPR} dataset. This dataset contains 10,594 CT studies from 4,427 unique patients. There are 32,735 lesions annotated at their key slices. The whole dataset is officially divided into training, validation, and testing set with each of them containing 22,901, 4,887, 4,912 lesions respectively (noisy annotations are removed). Note that, DeepLesion only provides 60mm Z-context along with the key slice for each lesion. On the other hand, various types of lesions are included in this dataset, including lung, mediastinum, liver, soft tissue, pelvis, abdomen, kidney, and bone. This wide variety of lesions allows us to evaluate our approach on a more general scale.

However, it is reported in \cite{yan20183d} that 3D CNN does not work well with the DeepLesion dataset. We attribute this observation to three aspects: (1) key slice indices may not be accurate especially when the slice interval is large (eg., 5mm); (2) large lesions ($\geq$ 48mm, $\sim$11\% of the data) can be easily out-of-bound; (3) annotated bounding boxes for small lesions are usually too large compared with the actual size of lesions. All these issues pose significant challenges to the bounding box regression. To attack these issues we merge multi-annotated lesions (lesions with multiple annotations), remove very large lesions during training and adjust the diameter of small lesions to the minimum of the long side of the bounding box and long diameter. 

\subsection{Data Pre-processing}
Note that no semantic segmentation is applied here because whole CT scans are not available. However, we still can reduce unnecessary parts by clipping black borders. Similar to the experiments on LUNA16, each scan chunk is rescaled to an isotropic resolution (1 mm). In all experiments with DeepLesion, training sample size is 64$\times$128$\times$128 (padding 0 when necessary).

We convert the 2D annotations into 3D ones as \{X, Y, Z, Diameter\} vectors. Z position is calculated by key slice indices and slice intervals. In this way, the task settings become the same as LUNA16. Moreover, despite the issues of bounding box regression, our model can still generalize well with DeepLesion. 

\subsection{Training and Testing Settings}
We adopt the same model architecture as with LUNA16 except for the anchor setting. The anchors on stride 4, 8 and 16 are configured as \{3, 5, 7\}, \{10, 13, 17\} and \{22.0, 30.0, 40.0\} respectively. In both training and testing, we adopt the same cropping strategy in LUNA16 experiments. We remove very large lesions ($\geq$ 48mm, $\sim$11\% of the data) during training. This operation is nontrivial. Our primary attempts show that when these large lesions are included, the regression losses from the RPN branch are hard to converge. When testing, we report both the results with and without very large nodules (also $\sim$11\% of the testing lesions). Note that, we adopt LUNA16's criterion for the evaluation. We train the model from scratch and results are summarized in Table \ref{tb:deeplesion}. 

\begin{table*}
\caption{Sensitivity (\%) and FROC score on the DeepLesion dataset. With ``*" means large lesions ($>$48 mm) are removed. Note that we may not directly compare performance with \cite{yan20183d} because of the different evaluation criteria (2D vs 3D).}
    \centering
    \begin{tabular}{c|c|c|c|c|c|c|c|c} 
    FPs per image & 0.5 & 1 & 2 & 4 & 8 & 16 & Avg. & FROC \\ \hline
    3DCE, 27 slices \cite{yan20183d} & 62.48 &73.37 &80.70 &85.65 &89.09 &91.06 & 80.39 & - \\\Xhline{3\arrayrulewidth} 
    RPN Baseline  & 65.74 &73.89 &80.99 &86.56 & 91.40 & 94.40 & 82.17 & 0.708 \\\hline
    3D-AG  & \textbf{74.08} &\textbf{81.42}  &\textbf{86.08}  &\textbf{89.38} 
    &\textbf{92.08}  &\textbf{94.82} & \textbf{86.31} & \textbf{0.771} \\\hline
    \Xhline{3\arrayrulewidth} 
    RPN Baseline * & 69.09 & 76.75 &82.81 &87.29 &91.20 &93.84 & 83.50 & 0.735 \\\hline
    
    3D-AG * &  \textbf{74.68}  &\textbf{81.52}  &\textbf{85.45}  &\textbf{89.18} &\textbf{92.01}  &\textbf{94.72} & \textbf{86.26} & \textbf{0.774} \\\hline
 \end{tabular}
\label{tb:deeplesion}   
\end{table*}

\begin{table*}
\caption{Sensitivity@4 (\%) on DeepLession w.r.t Lesion Type and Diameter. With ``*" means large lesions ($>$48 mm) are removed. The abbreviations of lesion types stand for lung (LU), mediastinum (ME), liver (LV), soft tissue (ST), pelvis (PV), abdomen (AB), kidney (KD), and bone (BN), respectively. ``$<$10", ``10-30" and ``$>$30" indicate lesion diameter ranges (mm).}
    \centering
    \begin{tabular}{c|cccccccc|ccc} 
    Model &LU &ME &LV &ST &PV &AB &KD &BN & $<$10 &10-30 & $>$30  \\ \hline
    3DCE, 27 slices \cite{yan20183d} & 89 &88 &90 &74 &84 &84 &82 &75 &80 &87 &84 \\\Xhline{3\arrayrulewidth} 
    RPN Baseline  & 91 &88 &87 &80 &85 &80 &80 &69 &82 &88 &80   \\\hline
    3D-AG  & \textbf{93} &\textbf{89} &\textbf{92} &\textbf{84} &\textbf{90} &\textbf{86} & \textbf{83} &\textbf{70} &\textbf{83} &\textbf{90} &\textbf{91} \\\Xhline{3\arrayrulewidth} 
    RPN Baseline * & 91 &88 &89 &80 &86 &82 &81 &68 &82 &88 &84  \\\hline
    3D-AG * & \textbf{93} &\textbf{89} &\textbf{91} &\textbf{84} &\textbf{90} &\textbf{86} &\textbf{83} &\textbf{70} &\textbf{83} &\textbf{90} &\textbf{90}  \\\hline
 \end{tabular}
\label{tb:deeplesion_subtype}
\end{table*}

\begin{figure*}
\begin{center}
\includegraphics[width=7in]{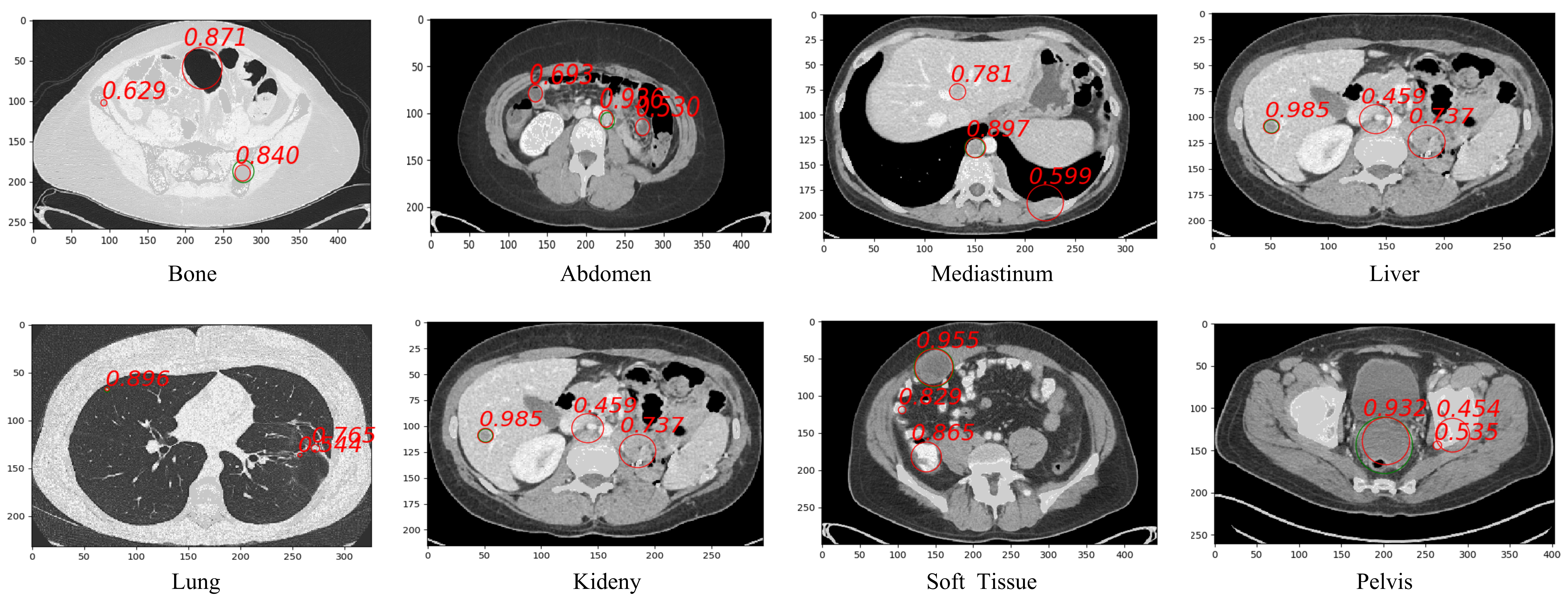}
\end{center}
\caption{Visualization with DeepLesion. Green boxes are ground truth while red boxes are ones predicted. It may be not clear sometimes because of the nearly perfect match. Such as Abdomen: box 0.936, Mediastinum: box 0.897, Liver: 0.985, Lung: box 0.896, Kidney: box 0.985, Soft Tissue: 0.865. }
\label{vis_DeepLesion}
\end{figure*} 

\subsection{Overall Performance}
We evaluate the performance with the FROC score here. As we can see from Table \ref{tb:deeplesion}, our model generalizes well to the general lesion detection task and the Aggregated FPRN branch consistently improves the baseline RPN. Again, the large performance gap between RPN w/ and w/o very large lesions supports our conjecture that our model could be sensitive to out-of-bound lesions. Nevertheless, it shows that our Aggregated FPRN is robust to the detection of a variety of lesions. Moreover, our Aggregated FPRN is less sensitive to the out-of-bound lesions. This is another advantage brought by our model. 

\subsection{Performance w.r.t. Lesion Type and Size}
As in \cite{yan20183d}, we also report the performance with respect to Lesion Type and Diameter. All results are summarized in Table \ref{tb:deeplesion_subtype}. From Table \ref{tb:deeplesion_subtype} we can easily find that our model does not perform well for bone and kidney lesions. On the other hand, our approach does not experience significant performance drop as in \cite{yan20183d} when detecting ``Soft Tissue" lesions. Again, we stress the point that one may not directly compare our results with \cite{yan20183d}. Nevertheless, the considerable improvement brought by the Aggregated classifier branch compared with the RPN baseline shows that our Aggregated Faster R-CNN generalizes well across different tasks. Some of the results are visualized in Fig. \ref{vis_DeepLesion}.

\section{Conclusions and Future work}
For the training of the RPN branch, the adaptive Focal Loss demonstrates superiority to the OHEM mechanism in terms of stability, training speed and performance. The boosting strategy Anchor Based Sampling proves to be effective. Moreover, we have observed that segmentation also plays an important role by filtering out noise and reducing the unnecessary area. 
Inspired by these observations, we can further improve the training mechanism in three directions: (1) the sampling strategy and robust segmentation at the starting point; (2) the backbone network structure in the middle; (3) the cost at the ending point. 

By stacking the FPRN branch RPNs' backbone, we can consistently improve the FROC performance on this lesion detection task. One surprising finding is that with the help of the FPRN branch, the model becomes more robust to out-of-bound lesions. When isolating each technical component, we find that (1) Diameter Alignment plays a critical role by enriching the context information; (2) Local Magnification Operations are effective for the FPRN branch. Sometimes, however, it may not be the best choice for the full solution. This inconsistency calls for further research. Possible directions can be better designed FPRN structure, losses balance, and local regional constraints. Nevertheless, this Local Magnification and FPRN branch open a door towards a full diagnosis of nodules by offering more interpretable features: texture, calcification, lobulation, and even malignancy. All these features can potentially be incorporated into the FPRN branch. We will focus on this interpretability in the future.

\section{Acknowledgment}

\bibliographystyle{IEEEtran}
\bibliography{egbib}

\begin{thebibliography}{10}
\providecommand{\url}[1]{#1}
\csname url@samestyle\endcsname
\providecommand{\newblock}{\relax}
\providecommand{\bibinfo}[2]{#2}
\providecommand{\BIBentrySTDinterwordspacing}{\spaceskip=0pt\relax}
\providecommand{\BIBentryALTinterwordstretchfactor}{4}
\providecommand{\BIBentryALTinterwordspacing}{\spaceskip=\fontdimen2\font plus
\BIBentryALTinterwordstretchfactor\fontdimen3\font minus
  \fontdimen4\font\relax}
\providecommand{\BIBforeignlanguage}[2]{{%
\expandafter\ifx\csname l@#1\endcsname\relax
\typeout{** WARNING: IEEEtran.bst: No hyphenation pattern has been}%
\typeout{** loaded for the language `#1'. Using the pattern for}%
\typeout{** the default language instead.}%
\else
\language=\csname l@#1\endcsname
\fi
#2}}
\providecommand{\BIBdecl}{\relax}
\BIBdecl

\bibitem{chen2019mini}
Q.~Chen, X.~Sun, N.~Zhang, Y.~Cao, and B.~Liu, ``Mini lesions detection on
  diabetic retinopathy images via large scale cnn features,'' in
  \emph{International Conference on Tools with Artificial Intelligence
  (ICTAI)}.\hskip 1em plus 0.5em minus 0.4em\relax IEEE, 2019.

\bibitem{bejnordi2017diagnostic}
B.~E. Bejnordi, M.~Veta, P.~J. Van~Diest, B.~Van~Ginneken, N.~Karssemeijer,
  G.~Litjens, J.~A. Van Der~Laak, M.~Hermsen, Q.~F. Manson, M.~Balkenhol
  \emph{et~al.}, ``Diagnostic assessment of deep learning algorithms for
  detection of lymph node metastases in women with breast cancer,''
  \emph{Jama}, vol. 318, no.~22, pp. 2199--2210, 2017.

\bibitem{setio2017validation}
A.~A.~A. Setio, A.~Traverso, T.~De~Bel, M.~S. Berens, C.~van~den Bogaard,
  P.~Cerello, H.~Chen, Q.~Dou, M.~E. Fantacci, B.~Geurts \emph{et~al.},
  ``Validation, comparison, and combination of algorithms for automatic
  detection of pulmonary nodules in computed tomography images: the luna16
  challenge,'' \emph{Medical image analysis}, vol.~42, pp. 1--13, 2017.

\bibitem{henschke2002ct}
C.~I. Henschke, D.~F. Yankelevitz, R.~Mirtcheva, G.~McGuinness, D.~McCauley,
  and O.~S. Miettinen, ``Ct screening for lung cancer: frequency and
  significance of part-solid and nonsolid nodules,'' \emph{American Journal of
  Roentgenology}, vol. 178, no.~5, pp. 1053--1057, 2002.

\bibitem{murphy2009large}
K.~Murphy, B.~van Ginneken, A.~M. Schilham, B.~De~Hoop, H.~Gietema, and
  M.~Prokop, ``A large-scale evaluation of automatic pulmonary nodule detection
  in chest ct using local image features and k-nearest-neighbour
  classification,'' \emph{Medical image analysis}, vol.~13, no.~5, pp.
  757--770, 2009.

\bibitem{jacobs2014automatic}
C.~Jacobs, E.~M. van Rikxoort, T.~Twellmann, E.~T. Scholten, P.~A. de~Jong,
  J.-M. Kuhnigk, M.~Oudkerk, H.~J. de~Koning, M.~Prokop, C.~Schaefer-Prokop
  \emph{et~al.}, ``Automatic detection of subsolid pulmonary nodules in
  thoracic computed tomography images,'' \emph{Medical image analysis},
  vol.~18, no.~2, pp. 374--384, 2014.

\bibitem{yan20183d}
K.~Yan, M.~Bagheri, and R.~M. Summers, ``3d context enhanced region-based
  convolutional neural network for end-to-end lesion detection,'' in
  \emph{International Conference on Medical Image Computing and
  Computer-Assisted Intervention}.\hskip 1em plus 0.5em minus 0.4em\relax
  Springer, 2018, pp. 511--519.

\bibitem{setio2016pulmonary}
A.~A.~A. Setio, F.~Ciompi, G.~Litjens, P.~Gerke, C.~Jacobs, S.~J. van Riel,
  M.~M.~W. Wille, M.~Naqibullah, C.~I. S{\'a}nchez, and B.~van Ginneken,
  ``Pulmonary nodule detection in ct images: false positive reduction using
  multi-view convolutional networks,'' \emph{IEEE transactions on medical
  imaging}, vol.~35, no.~5, pp. 1160--1169, 2016.

\bibitem{liao2017evaluate}
F.~Liao, M.~Liang, Z.~Li, X.~Hu, and S.~Song, ``Evaluate the malignancy of
  pulmonary nodules using the 3d deep leaky noisy-or network,'' \emph{arXiv
  preprint arXiv:1711.08324}, 2017.

\bibitem{zhu2018deeplung}
W.~Zhu, C.~Liu, W.~Fan, and X.~Xie, ``Deeplung: Deep 3d dual path nets for
  automated pulmonary nodule detection and classification,'' \emph{arXiv
  preprint arXiv:1801.09555}, 2018.

\bibitem{zhongliuxie2018}
Z.~Xie, ``Towards single-phase single-stage detection of pulmonary nodules in
  chest ct imaging,'' \emph{arXiv preprint arXiv:1807.05972}, 2018.

\bibitem{ren2015faster}
S.~Ren, K.~He, R.~Girshick, and J.~Sun, ``Faster r-cnn: Towards real-time
  object detection with region proposal networks,'' in \emph{Advances in neural
  information processing systems}, 2015, pp. 91--99.

\bibitem{mahmud2018applications}
M.~Mahmud, M.~S. Kaiser, A.~Hussain, and S.~Vassanelli, ``Applications of deep
  learning and reinforcement learning to biological data,'' \emph{IEEE
  transactions on neural networks and learning systems}, vol.~29, no.~6, pp.
  2063--2079, 2018.

\bibitem{fu2017dssd}
C.-Y. Fu, W.~Liu, A.~Ranga, A.~Tyagi, and A.~C. Berg, ``Dssd: Deconvolutional
  single shot detector,'' \emph{arXiv preprint arXiv:1701.06659}, 2017.

\bibitem{he2017mask}
K.~He, G.~Gkioxari, P.~Doll{\'a}r, and R.~Girshick, ``Mask r-cnn,'' in
  \emph{Computer Vision (ICCV), 2017 IEEE International Conference on}.\hskip
  1em plus 0.5em minus 0.4em\relax IEEE, 2017, pp. 2980--2988.

\bibitem{ronneberger2015u}
O.~Ronneberger, P.~Fischer, and T.~Brox, ``U-net: Convolutional networks for
  biomedical image segmentation,'' in \emph{International Conference on Medical
  image computing and computer-assisted intervention}.\hskip 1em plus 0.5em
  minus 0.4em\relax Springer, 2015, pp. 234--241.

\bibitem{long2015fully}
J.~Long, E.~Shelhamer, and T.~Darrell, ``Fully convolutional networks for
  semantic segmentation,'' in \emph{Proceedings of the IEEE conference on
  computer vision and pattern recognition}, 2015, pp. 3431--3440.

\bibitem{he2016deep}
K.~He, X.~Zhang, S.~Ren, and J.~Sun, ``Deep residual learning for image
  recognition,'' in \emph{Proceedings of the IEEE conference on computer vision
  and pattern recognition}, 2016, pp. 770--778.

\bibitem{Zhang:2017:IMR:3123266.3123332}
\BIBentryALTinterwordspacing
N.~Zhang, Y.~Cao, B.~Liu, and Y.~Luo, ``Improved multimodal representation
  learning with skip connections,'' in \emph{Proceedings of the 2017 ACM on
  Multimedia Conference}, ser. MM '17.\hskip 1em plus 0.5em minus 0.4em\relax
  New York, NY, USA: ACM, 2017, pp. 654--662. [Online]. Available:
  \url{http://doi.acm.org/10.1145/3123266.3123332}
\BIBentrySTDinterwordspacing

\bibitem{he2016identity}
K.~He, X.~Zhang, S.~Ren, and J.~Sun, ``Identity mappings in deep residual
  networks,'' in \emph{European conference on computer vision}.\hskip 1em plus
  0.5em minus 0.4em\relax Springer, 2016, pp. 630--645.

\bibitem{lin2017feature}
T.-Y. Lin, P.~Doll{\'a}r, R.~Girshick, K.~He, B.~Hariharan, and S.~Belongie,
  ``Feature pyramid networks for object detection,'' in \emph{CVPR}, vol.~1,
  no.~2, 2017, p.~4.

\bibitem{setio2015automatic}
A.~A. Setio, C.~Jacobs, J.~Gelderblom, and B.~Ginneken, ``Automatic detection
  of large pulmonary solid nodules in thoracic ct images,'' \emph{Medical
  physics}, vol.~42, no.~10, pp. 5642--5653, 2015.

\bibitem{tan2011novel}
M.~Tan, R.~Deklerck, B.~Jansen, M.~Bister, and J.~Cornelis, ``A novel
  computer-aided lung nodule detection system for ct images,'' \emph{Medical
  physics}, vol.~38, no.~10, pp. 5630--5645, 2011.

\bibitem{DBLP:journals/corr/DingLHW17}
\BIBentryALTinterwordspacing
J.~Ding, A.~Li, Z.~Hu, and L.~Wang, ``Accurate pulmonary nodule detection in
  computed tomography images using deep convolutional neural networks,''
  \emph{CoRR}, vol. abs/1706.04303, 2017. [Online]. Available:
  \url{http://arxiv.org/abs/1706.04303}
\BIBentrySTDinterwordspacing

\bibitem{ypsilantis2016recurrent}
P.-P. Ypsilantis and G.~Montana, ``Recurrent convolutional networks for
  pulmonary nodule detection in ct imaging,'' \emph{arXiv preprint
  arXiv:1609.09143}, 2016.

\bibitem{ioffe2015batch}
S.~Ioffe and C.~Szegedy, ``Batch normalization: Accelerating deep network
  training by reducing internal covariate shift,'' \emph{arXiv preprint
  arXiv:1502.03167}, 2015.

\bibitem{shrivastava2016training}
A.~Shrivastava, A.~Gupta, and R.~Girshick, ``Training region-based object
  detectors with online hard example mining,'' in \emph{Proceedings of the IEEE
  Conference on Computer Vision and Pattern Recognition}, 2016, pp. 761--769.

\bibitem{lin2017focal}
T.-Y. Lin, P.~Goyal, R.~Girshick, K.~He, and P.~Doll{\'a}r, ``Focal loss for
  dense object detection,'' \emph{arXiv preprint arXiv:1708.02002}, 2017.

\bibitem{dou2017multilevel}
Q.~Dou, H.~Chen, L.~Yu, J.~Qin, and P.-A. Heng, ``Multilevel contextual 3-d
  cnns for false positive reduction in pulmonary nodule detection,'' \emph{IEEE
  Transactions on Biomedical Engineering}, vol.~64, no.~7, pp. 1558--1567,
  2017.

\bibitem{zagoruyko2016wide}
S.~Zagoruyko and N.~Komodakis, ``Wide residual networks,'' \emph{arXiv preprint
  arXiv:1605.07146}, 2016.

\bibitem{Yan_2018_CVPR}
K.~Yan, X.~Wang, L.~Lu, L.~Zhang, A.~P. Harrison, M.~Bagheri, and R.~M.
  Summers, ``Deep lesion graphs in the wild: Relationship learning and
  organization of significant radiology image findings in a diverse large-scale
  lesion database,'' in \emph{The IEEE Conference on Computer Vision and
  Pattern Recognition (CVPR)}, June 2018.

\bibitem{huang2017densely}
G.~Huang, Z.~Liu, K.~Q. Weinberger, and L.~van~der Maaten, ``Densely connected
  convolutional networks,'' in \emph{Proceedings of the IEEE conference on
  computer vision and pattern recognition}, vol.~1, no.~2, 2017, p.~3.

\bibitem{lin2014microsoft}
T.-Y. Lin, M.~Maire, S.~Belongie, J.~Hays, P.~Perona, D.~Ramanan,
  P.~Doll{\'a}r, and C.~L. Zitnick, ``Microsoft coco: Common objects in
  context,'' in \emph{Computer Vision--ECCV 2014}.\hskip 1em plus 0.5em minus
  0.4em\relax Springer, 2014, pp. 740--755.

\bibitem{armato2011lung}
S.~G. Armato~III, G.~McLennan, L.~Bidaut, M.~F. McNitt-Gray, C.~R. Meyer, A.~P.
  Reeves, B.~Zhao, D.~R. Aberle, C.~I. Henschke, E.~A. Hoffman \emph{et~al.},
  ``The lung image database consortium (lidc) and image database resource
  initiative (idri): a completed reference database of lung nodules on ct
  scans,'' \emph{Medical physics}, vol.~38, no.~2, pp. 915--931, 2011.

\bibitem{eggert2017closer}
C.~Eggert, S.~Brehm, A.~Winschel, D.~Zecha, and R.~Lienhart, ``A closer look:
  Small object detection in faster r-cnn,'' in \emph{Multimedia and Expo
  (ICME), 2017 IEEE International Conference on}.\hskip 1em plus 0.5em minus
  0.4em\relax IEEE, 2017, pp. 421--426.

\end{thebibliography}

\end{document}